\documentclass[journal]{IEEEtran}

\usepackage{todonotes}

\usepackage{amsmath,amsfonts}
\usepackage{amsthm}
\usepackage[ruled,vlined]{algorithm2e}
\usepackage{algpseudocode}
\usepackage{array}
\usepackage[caption=false,font=normalsize,labelfont=sf,textfont=sf]{subfig}
\usepackage{textcomp}
\usepackage{stfloats}
\usepackage{url}
\usepackage{verbatim}
\usepackage{graphicx}
\usepackage{enumitem}
\usepackage{color}
\usepackage{cite}

\begin{document}

\title{Generating Synthetic Net Load Data with Physics-informed Diffusion Model}

\author{Shaorong Zhang,~\IEEEmembership{Student Member,~IEEE}, Yuanbin Cheng,~\IEEEmembership{Student Member,~IEEE}, Nanpeng Yu,~\IEEEmembership{Senior Member,~IEEE}
}



\maketitle

\begin{abstract}

This paper presents a novel physics-informed diffusion model for generating synthetic net load data, addressing the challenges of data scarcity and privacy concerns. The proposed framework embeds physical models within denoising networks, offering a versatile approach that can be readily generalized to unforeseen scenarios. A conditional denoising neural network is designed to jointly train the parameters of the transition kernel of the diffusion model and the parameters of the physics-informed function. Utilizing the real-world smart meter data from Pecan Street, we validate the proposed method and conduct a thorough numerical study comparing its performance with state-of-the-art generative models, including generative adversarial networks, variational autoencoders, normalizing flows, and a well calibrated baseline diffusion model. A comprehensive set of evaluation metrics is used to assess the accuracy and diversity of the generated synthetic net load data. The numerical study results demonstrate that the proposed physics-informed diffusion model outperforms state-of-the-art models across all quantitative metrics, yielding at least 20\% improvement.

\end{abstract}


\section{Introduction}



Having access to energy consumption data at the customer level is crucial to the development of distribution system operation and planning tools \cite{silva2023generating}. As advanced metering infrastructure expands, it provides a wealth of net load measurements that are instrumental for informed decision-making \cite{zhang2019scenario}, promoting efficient and reliable operation of the distribution system. Nonetheless, it is still challenging for some electric utilities to obtain net load data for all customers due to the significant costs for the installation and maintenance of smart meters \cite{dehdarian2018scenario}. Furthermore, industry developers and academic researchers often struggle to acquire real-world smart meter data due to privacy and security concerns \cite{lee2021data, lee2020federated}. Even if the access to smart meter data is granted, the availability of data under extreme operating conditions are often very limited.

To address these challenges, generating synthetic net-load data \cite{silva2023generating} has emerged as a promising approach to provide realistic energy consumption data for research and development purposes. By preserving the essential spatio-temporal correlations of real-world data \cite{yilmaz2022synthetic}, synthetic net-load data becomes a cornerstone for many power system studies such as
power flow analysis, stability assessments, fault analysis, and demand-side management \cite{pinceti2019data, menati2023modeling}. Synthetic datasets are not only instrumental in representing a wide variety of operating scenarios for these studies but also crucial for multi-period economic dispatch, unit commitment, system planning, and long-term reliability assessments \cite{asare2014future, pinto2022transfer}. Electric load data synthesis \cite{wang2023customized} is also an effective technique that provides ample training data for developing data-driven and machine learning algorithms in power distribution systems.

To generate synthetic load data, a variety of methods have been developed and refined over time. Model-based methods involve creating physical models that can generate synthetic load curves. They use historical data to capture power consumption patterns and then learn physical properties in different scenarios. For example, residential energy consumption can be simulated by modeling household appliances and physical characteristics of buildings \cite{diao2017modeling, lopez2018smart}. However, model-based methods require detailed physics-based models and accurate parameters of the physical systems, making them difﬁcult to adopt and generalize across different scenarios \cite{wang2023customized}. 

To solve the above problems, a growing body of research on load synthesis is turning towards data-driven methods, such as principal component analysis \cite{pinceti2019data}, nonlinear independent component estimation \cite{silva2023generating}, probabilistic models \cite{dickert2011time, gruber2012residential}, autoregressive models and Markov chains \cite{liu2019two}. 
Researchers also used clustering techniques such as K-means \cite{al2017k} and fuzzy c-means clustering \cite{kim2011study} to segment load profiles into different groups, which represent various customer categories. The clustering techniques can be combined with other data-driven methods such as Markov models to synthesize residential loads in a top-down manner \cite{6412795}.

The data-driven methods have evolved with the advancement of machine learning algorithms, with earlier works focusing on end-to-end learning by neural networks. In \cite{pillai2014generation}, an artificial neural network (ANN) based method is proposed to synthesize load profiles for a target region using its weather data as inputs. In \cite{sarochar2019synthesizing}, mixture density network (MDN) model was integrated with a multi-layered long short-term memory (LSTM) network to synthesize energy consumption data. In \cite{salazar2020data}, a framework that combined transfer learning with the domain adaptation approach was developed for load profile generation in medium voltage networks.

In recent years, researchers who study synthetic load data generation embraced sophisticated generative models, such as variational auto-encoders (VAEs), Normalizing Flows (NFs), and generative adversarial networks (GANs). For instance, VAEs have been effectively utilized for generating electric vehicle (EV) load profiles, as demonstrated in \cite{pan2019data}. While conditional VAEs have been tailored to generate contextual load profiles based on temporal conditions and grid interactions in \cite{wang2022generating}. The VAEs suﬀer from inherent shortcomings, such as the difﬁculties of tuning hyper-parameters or generalizing a speciﬁc generative model structure to other databases \cite{dumas2022deep}.

GANs have also been employed to create synthetic load patterns and energy usage behaviors \cite{el2020data} and realistic building electric load profiles \cite{wang2020generating}, and predict daily load profiles \cite{bendaoud2021comparing}. GANs have proven to be quite capable of synthesizing fine-grained energy consumption time series \cite{li2022energy}, and recovering high-resolution load profiles from low-resolution ones \cite{song2022profilesr}. More advanced GAN models have also been used to generate synthetic load data. To preserve customers’ data privacy, differentially private Wasserstein generative adversarial networks (DPWGAN) were developed to synthesize high-quality load proﬁles \cite{huang2022dpwgan}. Considering spatial-temporal correlations, multi-load generative adversarial network (MultiLoad-GAN) was proposed to generate a group of synthetic load profiles simultaneously.

With the increasing integration of intermittent energy resources, such as rooftop solar photovoltaics (PVs), EVs and demand response programs, the pattern of the net load becomes more complex, which increases the difficulty of the net load synthesis task. However, most of the work using GAN models to create synthetic load data did not beneﬁt from conditional information such as weather forecasts and solar PV capacity. This is because GAN can be tailored to condition on discrete variables but not on continuous and vector-valued ones in the synthetic load data generation task. Furthermore, the training of GAN models is notoriously unstable because of the two constantly competing components: the generator and the discriminator \cite{arjovsky2017wasserstein}. 

To use continuous variables as conditional information, some researchers start to use the NF model to create synthetic load data \cite{chen2018model, zhang2019scenario}.
However, the generalizability of ﬂow-based models is limited because of their reliance on specialized architectures, which must be individually designed for each case to establish  reversible transformations \cite{wang2023customized}. 

Diffusion models overcome several key challenges in other generative models: the problem of aligning posterior distributions in VAEs, the unstable adversarial objective in GANs, the long training-time of Markov Chain Monte Carlo (MCMC) methods in energy-based models (EBMs), and imposing network constraints as in normalizing flows \cite{cao2022survey}. Diffusion models have showcased tremendous success on many applications such as image generation \cite{ho2020denoising} and audio generation \cite{chen2020wavegrad}.

Addressing the limitations of GAN, VAE and NF, diffusion models have emerged as the leading choice for generative models. Their applications extend well beyond image synthesis. In recent years, diffusion models have been effectively utilized in various time series generation tasks. These include  generation of sleep electroencephalography signals \cite{aristimunha2023synthetic}, 
wind power scenario generation \cite{dong2023short}, and EV charging scenario generation \cite{li2024diffcharge}. Demonstrating significant advantages over other generative models, diffusion models excel at capturing the complex statistical properties and temporal dynamics inherent in time series data.

Therefore, we adapt diffusion models to generate net load data in this paper. In order to fully utilize physical models, we propose a physics-informed diffusion model (PDM) by embeding solar PV System Performance Model (PVSPM) into the baseline diffusion model (BDM). We propose to jointly estimate the parameters of physics-informed function and the transition kernel of the BDM in the tailored conditional denoising network. We evaluate the proposed PDM with state-of-the-art generative models such as BDM, GAN, VAE, NF using publicly available net-load data from Pecan Street \cite{PecanStreet2020}.

The main contributions of this paper are listed below:

\begin{itemize}
    \item We propose a physics-informed diffusion framework for net load data synthesis. This framework integrates physical solar PV performance model directly into the denoising network, making it more interpretable and capable of generalizing to unforeseen conditions.
    \item To capture the temporal correlations of the net load profiles and fully utilize the physical model, we design both the baseline and the physics-informed denoising networks. These networks effectively combine and integrate LSTM units, multi-head self-attention mechanisms, multi-layer perceptrons, and physical models, among other components, to optimize the model performance.
    \item The superior performance of the proposed method is verified through comprehensive numerical studies. We compare the performance of our proposed PDM with an extensive list of generative models, including GAN, VAE, NF and BDM on the net load synthetic task. It is shown that the proposed PDM achieves over 20\% improvement over other methods across all evaluation metrics.
\end{itemize}

The remainder of this paper is organized as follows. Section \ref{sec:Prob_formu} formulates the net load synthesis problem. 
Section \ref{sec: technical_methods} introduces the proposed technical methods. Section \ref{sec: nenumerical_studies} presents the numerical studies. Section \ref{sec:conclusion} gives the conclusion.

\section{Problem Formulation}
\label{sec:Prob_formu}
The aim of this paper is to generate the net load data of residential customers with solar PV systems. The net load readings can be decomposed into the solar PV generation and electric load consumption. According to the net load deﬁnition, the net load, electric load, and solar generation of a customer satisfy the following equality constraint:
\begin{equation}
    \mathbf{NL} =  \mathbf{Lo} -  \mathbf{So},
\end{equation}
where $ \mathbf{NL} \in \mathbb{R}^T$ denotes the net load measurements for a residential customer across a day, which is also referred to as net load profile in this paper. Here, $\mathbf{Lo} \in \mathbb{R}^T$ and $\mathbf{So} \in \mathbb{R}^T$ represent the electric load consumption and solar generation, respectively, each over the time horizon of one day.

\begin{figure}
    \centering
    \includegraphics[width = 0.5 \textwidth]{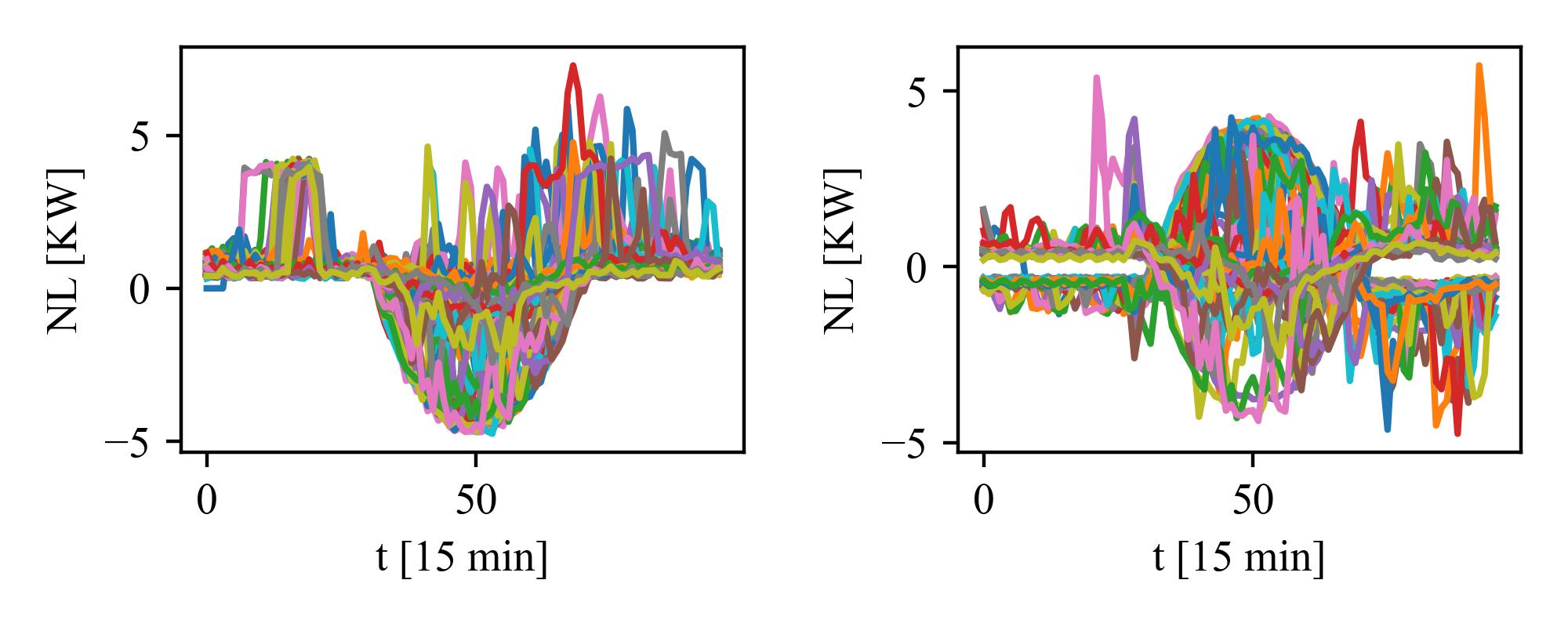}
    \caption{Net load profiles for two distinct customers in the first two months of 2018, with curves of different colors representing various days.}
    \label{fig:load_compare}
\end{figure}

In general, net load profiles differ daily and vary among customers. Fig. \ref{fig:load_compare} shows the net load profiles for two different customers in the first two months of 2018. The data sampling frequency is 15 minutes and the net-load readings are obtained from the Pecan street dataset \cite{PecanStreet2020}. The net load curves exhibit significant variations due to changing weather conditions, customer electricity consumption behaviors, and solar PV system configurations. Furthermore, the daily net load curves of a specific customer exhibit significant variations across time. 

Our task is to generate net load profiles utilizing customer ID, static solar PV system information, and other variables associated with the date. This task calls for the development of a conditional generative model, which is more challenging than designing a basic generative model without using context information. Let $\mathbf{y} \in \mathbb{R}^C $ represent the conditional information required for data generation, $\mathbf{y} = [\mathbf{y}_{ID}, \mathbf{y}_{PV}, \mathbf{y}_{D}]^T$, where $\mathbf{y}_{ID}\in \mathbb{R}^{C_{1}}$, $\mathbf{y}_{PV} \in \mathbb{R}^{C_{2}}$, $\mathbf{y}_{D} \in \mathbb{R}^{C_{3}}$ represent the conditional information that encoded user ID, solar PV system information, and variables associated the date, respectively. The solar PV system information includes the size of solar PV systems oriented towards the south, west, and east. The variables associated with specific dates are represented using one-hot encoding, which includes 12 dimensions for the month, 31 dimensions for the date within a month, and 7 dimensions for the day of the week. To generate net load data in specific scenarios, we need to learn the conditional distribution $p( \mathbf{NL} \vert \mathbf{y})$ using generative models. Note that it is quite challenging to learn the generative model for net load data generation, because $\mathbf{y}$ contains both continuous and discrete variables.

\section{Technical methods}
\label{sec: technical_methods}
\subsection{Denoising Diffusion Probabilistic Model for Net Load Profile Generation}

In this work, we build on top of the denoising diffusion probabilistic model (DDPM), which achieves state of the art results in  fields such as image generation \cite{ho2020denoising} and speech synthesis \cite{chen2020wavegrad}. Aimed at producing higher quality time series data, our implementation of the DDPM primarily draws inspiration from the speech synthesis model described in \cite{chen2020wavegrad}. The key modification is the utilization of the $L_2$ norm as loss function, which is more suitable for net load profile generation based on numerical study results.

\begin{figure*}[ht]
    \centering
    \includegraphics[width = 0.7\textwidth]{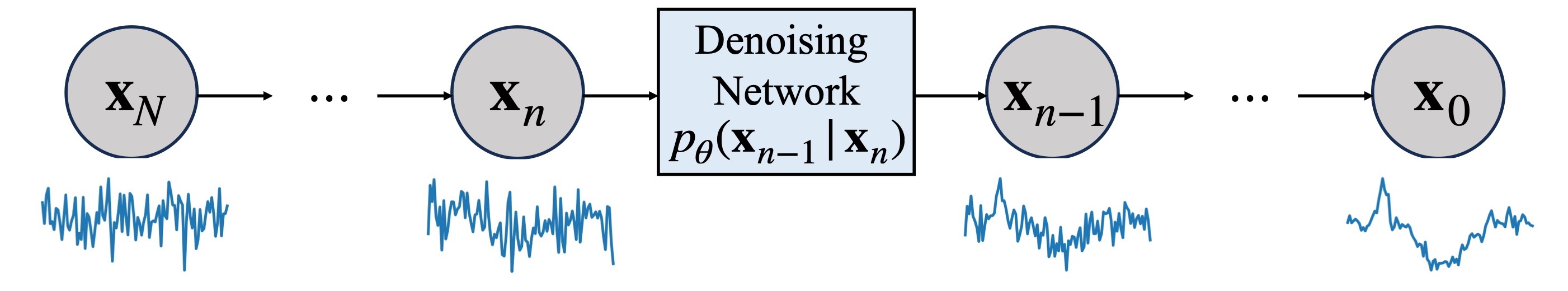}
    \caption{The overall framework of diffusion process for net load profile generation. The reverse denoising process progressively removes noise by denoising network $p_{\theta} (\mathbf{x}_{n-1} \vert \mathbf{x}_n)$, starting from $\mathbf{x}_N \sim \mathcal{N}(0, \mathbf{I})$, and concluding with $\mathbf{x}_0$.}
    \label{fig:diff_process}
\end{figure*}

Diffusion in the context of statistics refers to transforming a complex data distribution $p_{\text{data}}$ to a simple prior distribution $p_{\text{prior}}$ on the same domain. In the DDPM, this process is called the forward process. Conversely, the reverse process transforms $p_{\text{prior}}$ to $p_{\text{data}}$. In practice, we let $p_{\text{prior}}$ be a Gaussian distribution. Both the forward and the reverse processes can be modeled as Markov Chains, and the reverse data transformation can be learned by a deep denoising neural network. The overall framework of the diffusion process for net load profile generation is shown in Fig. \ref{fig:diff_process}.

\subsubsection{Forward Process}
The diffusion forward process in DDPM utilizes a Markov chain $\mathbf{x}_0, \mathbf{x}_1, ..., \mathbf{x}_N$ to progressively transform the data distribution $p_{\text{data}}$ into a predefined prior distribution, specifically a standard Gaussian distribution $\mathcal{N}(0, \mathbf{I})$. This transformation involves a series of steps where each transition from $\mathbf{x}_{t-1}$ to $\mathbf{x}_t$ is governed by a Gaussian distribution:
\begin{equation}
q\left(\mathbf{x}_n \vert \mathbf{x}_{n-1}\right):=\mathcal{N}\left(\mathbf{x}_n ; \sqrt{\left(1-\beta_n\right)} \mathbf{x}_{n-1}, \beta_n \mathbf{I}\right).
\end{equation}

Under some fixed noise schedule $\beta_1, \cdots, \beta_N$, the diffusion forward process with $n$ steps can be implemented using a closed form solution:
\begin{equation}
    \mathbf{x}_n = \sqrt{\bar{\alpha}_n} \mathbf{x}_0 + \sqrt{(1 - \bar{\alpha}_n )} \boldsymbol{\epsilon},
\end{equation}
where $\boldsymbol{\epsilon} \sim \mathcal{N} (0, I)$,  $\alpha_n := 1 - \beta_n$ and $\bar{\alpha}_n:=\prod_{s=1}^n \alpha_s$.

\subsubsection{Reverse Process}
The reverse process employs a Markov chain to transition from the prior distribution $p_{\text{prior}}$ back to the original data distribution $p_{\text{data}}$, effectively ``denoising" the data. The reverse process still follows a Gaussian distribution but lacks a closed-form transition kernel. We can use neural networks to parameterize the transition kernel:
\begin{equation}
    p_{\theta} (\mathbf{x}_{n-1} \vert \mathbf{x}_n) = \mathcal{N}(\mathbf{x}_{n-1}, \boldsymbol{\mu}_{\theta}(\mathbf{x}_n, n), \boldsymbol{\Sigma}_{\theta}(\mathbf{x}_n, n)).
\end{equation}

Instead of parameterizing both the mean $\boldsymbol{\mu}_{\theta}(\mathbf{x}_n, n)$ and covariance matrix $\boldsymbol{\Sigma}_{\theta}(\mathbf{x}_n, n))$ directly, we make the covariance matrix to be independent of $\mathbf{x}_n$, i.e., $\boldsymbol{\Sigma}_{\theta}(\mathbf{x}_n, n)) = \sigma_n^2 \mathbf{I}$, by following \cite{ho2020denoising}. Furthermore, we reparameterize the model to condition on the continuous noise level $\bar{\alpha}$ instead of the discrete iteration index $n$, which is shown to have a better performance in the audio generation task \cite{chen2020wavegrad}. Specifically, $p_{\theta} (\mathbf{x}_{n-1} \vert \mathbf{x}_n) = \mathcal{N}(\mathbf{x}_{n-1}, \boldsymbol{\mu}_{\theta}(x_n, \sqrt{\bar{\alpha}_n}, \mathbf{y}), \sigma_n^2 \mathbf{I})$, where $\mathbf{y}$ contains the conditioning features, $\boldsymbol{\mu}_{\theta}(\mathbf{x}_n, \sqrt{\bar{\alpha}_n}, \mathbf{y})$ is reparameterized as:
\begin{equation}
    \mu_{\theta}(\mathbf{x}_n, \sqrt{\bar{\alpha}_n}, \mathbf{y}) = \frac{1}{\sqrt{\alpha_n}} (\mathbf{x}_n - \frac{1 - \alpha_n}{\sqrt{1 - \bar{\alpha}_n}} \boldsymbol{\epsilon}_{\theta}(\mathbf{x}_n, \sqrt{\bar{\alpha}_n}, \mathbf{y})).
\end{equation}

The denoising network $\boldsymbol{\epsilon}_{\theta} (\mathbf{x}_n, \sqrt{\bar{\alpha}_n}, \mathbf{y}))$ can be trained by denoising score matching \cite{ho2020denoising}. The training objective is:
\begin{equation}
\label{objective}
\mathbb{E}_{n, \boldsymbol{\epsilon}} [\Vert \boldsymbol{\epsilon}_{\theta} (\sqrt{\bar{\alpha}_n} \mathbf{x}_0 + \sqrt{1 - \bar{\alpha}_n} \boldsymbol{\epsilon}, \sqrt{\bar{\alpha}_n}, \mathbf{y}) - \boldsymbol{\epsilon} \Vert_2],
\end{equation}
where $n \sim \text{Uniform} \{2, \cdots, N \}$ and $\boldsymbol{\epsilon} \sim \mathcal{N}(0, I)$.

After training the model, the sampling can be implemented using the Langevin dynamics:
\begin{equation}
\label{mchain}
    \mathbf{x}_{n-1} = \frac{1}{\sqrt{\alpha_n}} (\mathbf{x}_n - \frac{1 - \alpha_n}{\sqrt{1 - \bar{\alpha}_n}} \epsilon_{\theta} (\mathbf{x}_n, \sqrt{\bar{\alpha}_n}, \mathbf{y})) + \sigma_n \mathbf{z},
\end{equation}
where $\sigma_n = {\beta_n(1 - \bar{\alpha}_{n-1})}/{(1 - \bar{\alpha}_n)} $ and $1 < n  \leq N$. The pseudo-code for model training and scenario generation are summarized in Algorithm \ref{algo:train} and Algorithm \ref{algo:sample}.

\begin{algorithm}[ht]
\SetAlgoLined
\DontPrintSemicolon
\KwData{initialized meural network $\boldsymbol{\epsilon}_{\theta}$, noise schedule $\{ \beta_1, \cdots, \beta_N \}$, dataset sampled from $p_{data}$.}
\KwResult{trained network $\boldsymbol{\epsilon}_{\theta}$.}

Calculate $\bar{\alpha}_1, \cdots, \bar{\alpha}_N $, where  $\bar{\alpha}_n=\prod_{s=1}^n (1 - \beta_s)$

 \Repeat{converged}{
Sample data from training dataset: $\mathbf{x}_0 \sim p_{\text{data}}$\;

$n \sim \text{Uniform}\{2, \ldots, N \}$\;

$\sqrt{\bar{\alpha}} \sim \text{Uniform}(\sqrt{\bar{\alpha}}_{n-1}, \sqrt{\bar{\alpha}}_n)$\;

$\boldsymbol{\epsilon} \sim \mathcal{N} (0, \mathbf{I})$\;

\text{Take gradient descent step using}\;

$\, \nabla_{\theta} \Vert \boldsymbol{\epsilon} - \boldsymbol{\epsilon}_{\theta} (\sqrt{\bar{\alpha}}\mathbf{x}_0 + \sqrt{1 - \bar{\alpha}}\boldsymbol{\epsilon}, \sqrt{\bar{\alpha}}, \mathbf{y}) \Vert_2$\;
}
\caption{Model Training.}
\label{algo:train}
\end{algorithm}

\begin{algorithm}[ht]
\SetAlgoLined
\DontPrintSemicolon
\KwData{trained network $\boldsymbol{\epsilon}_{\theta}$, noise schedule $\{ \beta_1, \cdots, \beta_N \}$.}
\KwResult{$\mathbf{x}_0$.}

$\mathbf{x}_N \sim \mathcal{N}(0, \mathbf{I})$\;

\For{$n = N, \ldots, 1$}{
  $\mathbf{z} \sim \mathcal{N}(0, \mathbf{I})$\;
  
  $\mathbf{x}_{n-1} = \frac{1}{\sqrt{\alpha_n}} \left( \mathbf{x}_n - \frac{1 - \alpha_n}{\sqrt{1 - \bar{\alpha}_n}} \epsilon_{\theta} (\mathbf{x}_n, \sqrt{\bar{\alpha}_n}, \mathbf{y}) \right)$\;
  
  \If{$n > 1$}{
    $\mathbf{x}_{n-1} = \mathbf{x}_{n-1} + \sigma_n \mathbf{z}$\;
  }
}
\caption{Data Sampling}
\label{algo:sample}
\end{algorithm}

\subsection{Solar PV Generation Basis Profiles}

This subsection presents how we create solar PV generation basis profiles, which will be later used in the proposed PDM. The solar PV generation depends on the weather and solar PV system specifications. By leveraging the technical parameters a solar PV system along with pertinent weather data, the output of a solar PV system can be accurately estimated using the PVSPM. Our research employs models from two key sources: the PV performance model of Sandia National Laboratory \cite{stein2012photovoltaic} and PVWatts from the National Renewable Energy Laboratory \cite{dobos2014pvwatts}. The technical parameters of a solar PV system include the system's DC rating ($P_{dc0}$) in kW, the tilt angle ($\theta_t$) and azimuth angle ($\theta_{az}$) of the solar PV array, the nominal efficiency of the inverter ($\eta_{nom}$), and the system's overall loss ($l$). Weather data encompasses temperature, wind speed, direct normal irradiance (DNI), diffuse horizontal irradiance (DHI), and global horizontal irradiance (GHI), collectively represented by $ \mathbf{W} \in \mathbb{R}^{5 \times T}$.

The following procedure is followed to create the solar PV generation basis profiles. We pick a few representative azimuth angles ($\theta_{az}$) and form a set $A = \{90^\circ, 120^\circ, 150^\circ, 180^\circ, 210^\circ, 240^\circ, 270^\circ \}$. The tilt angle ($\theta_t$) is set to equal to the latitude of the customer location. The representative parameters of the solar PV system including the nominal inverter efficiency and panel loss are obtained from the Sandia Module database \cite{NREL_SAM}.
In the dataset, all customers are located in the same city with the same longitude and latitude for solar PV systems. Thus, for a given day, weather data $\mathbf{W}$ can be accurately determined by using the conditional information: $\mathbf{W} = \mathbf{W}(\mathbf{y}_{D})$. Consequently, variations in the solar PV generation basis profiles are attributed solely to the system size and azimuth angle. The solar PV output, given the specific date information $\mathbf{y}_{D}$ and azimuth angle $\theta_{az}$, can be calculated as follows:
\begin{equation}
    \mathbf{P}_{basis}(\theta_{az}, \mathbf{y}_{D}) = f_{\mathrm{PVSPM}} (W(\mathbf{y}_{D}), \theta_{az}),
\end{equation}
where $f_{\mathrm{PVSPM}}(\cdot)$ represents the PVSPM model, which uses the weather data and the azimuth angle as inputs. The output, $\mathbf{P}_{basis} (\theta_{az}, \mathbf{y}_{D}) \in \mathbb{R}^{T}$ serves as one of the solar PV generation basis profile associated with the azimuth angle $\theta_{az}$ and date $\mathbf{y}_{D}$.

The entire solar PV generation basis profiles for all $\theta_{az} \in A$ can be calculated:
\begin{equation}
    \mathbf{P}_{basis}(\mathbf{y}_{D}) = [\mathbf{P}_{basis}(90^{\circ}, \mathbf{y}_{D}), \cdots, \mathbf{P}_{basis}(270^{\circ}, \mathbf{y}_{D})]^T,
\end{equation}
where $\mathbf{P}_{basis} (\mathbf{y}_D) \in \mathbb{R}^{L \times T}$ is the entire solar PV generation basis under the date condition $\mathbf{y}_D$. $L = 7$ denotes the cardinality of set $A$, indicating the number of azimuth angles in the solar PV generation basis profiles. 


\subsection{Physics-informed Diffusion Framework}
Although diffusion models are successful in generating images and audio, they struggle to produce high quality net-load time series data. The net-load data can be partitioned into two distinct components: the electric load and solar PV generation. The second component can be calculated by a physical equation using the PVSPM. To effectively utilize the physical model, we propose a physics-informed diffusion framework. This innovative framework integrates physical models into the diffusion process, enhancing its generalizability and accuracy.

Let us assume that a signal $\mathbf{S}$ to be generated can be decomposed into two main components:
\begin{equation}
\label{decomp}
    \mathbf{S} = \mathbf{S}_d + \mathbf{S}_p,
\end{equation}
where $\mathbf{S}_d$ is the component that follows an unknown distribution, which will be learned by diffusion models. Thus, $\mathbf{S}_d = \mathbf{x}_0$ is calculated through a Markov chain $ \{\mathbf{x}_N, \cdots, \mathbf{x}_0 \}$, given $\mathbf{x}_N \sim \mathcal{N} (0, I)$ and transition kernel (\ref{mchain}). $\mathbf{S}_p$, on the other hand, can be calculated using a parameterized function with the basis profiles of a physical model and contextual variables. 

Suppose the physical component can be represented as:
\begin{equation}
    \mathbf{S}_p = f_{\phi}(\mathbf{y}, \mathbf{P}_{basis}(\mathbf{y}_D)),
    \label{phy_func}
\end{equation}
where $\mathbf{y}$ contains the contextual features and $\phi$ denotes the unknown parameters. $\mathbf{P}_{basis}$ is the PV generation basis profiles, which can be calculated through the PVSPM model. The details about $\mathbf{P}_{basis}$ will be introduced in the next subsection.

The first component of the time series, $\mathbf{S}_d$, is the end point of a Markov chain, whereas the second component $\mathbf{S}_p$ can be calculated through a parameterized physics-informed function. After both the Markov chain and the physics-informed function are learned, computing $\mathbf{S}$ becomes straightforward. However, since both the transition kernel and the physics-informed function contains unknown parameters, i.e., $\theta$ and $\phi$, it is difficult to learn them together.

The intuitive strategy might involve alternating updates for the two sets of parameters. However, this approach requires long training time and encounters instability and convergence issues. To address these challenges, we propose to learn $\theta$ and $\phi$ simultaneously by embedding the solar PV generation basis profiles $\mathbf{P}_{basis} (\mathbf{y}_D)$ into the denoising network. The details of this approach will be elaborated in the following subsection.

\subsection{Denoising Networks Design}
The most popular denoising network architecture in the image generation field is the U-net, which achieves state-of-the-art performance due to its ability to capture the spatial correlations of pixels in images. However, when it comes to generating net load profiles, the denoising network should be tailored to learn the temporal correlations and unique daily patterns in the energy usage. For instance, residential energy consumption profile often peaks during morning and early evening hours, which coincide with the daily routines of residential customers. Similarly, solar generation profiles are closely linked to the intensity of solar radiation and has strong temporal correlations. In this subsection, we design a denoising network that is tailored for synthesizing net load time series data.

\begin{figure}[ht]
    \centering
    \includegraphics[width = 0.45\textwidth]{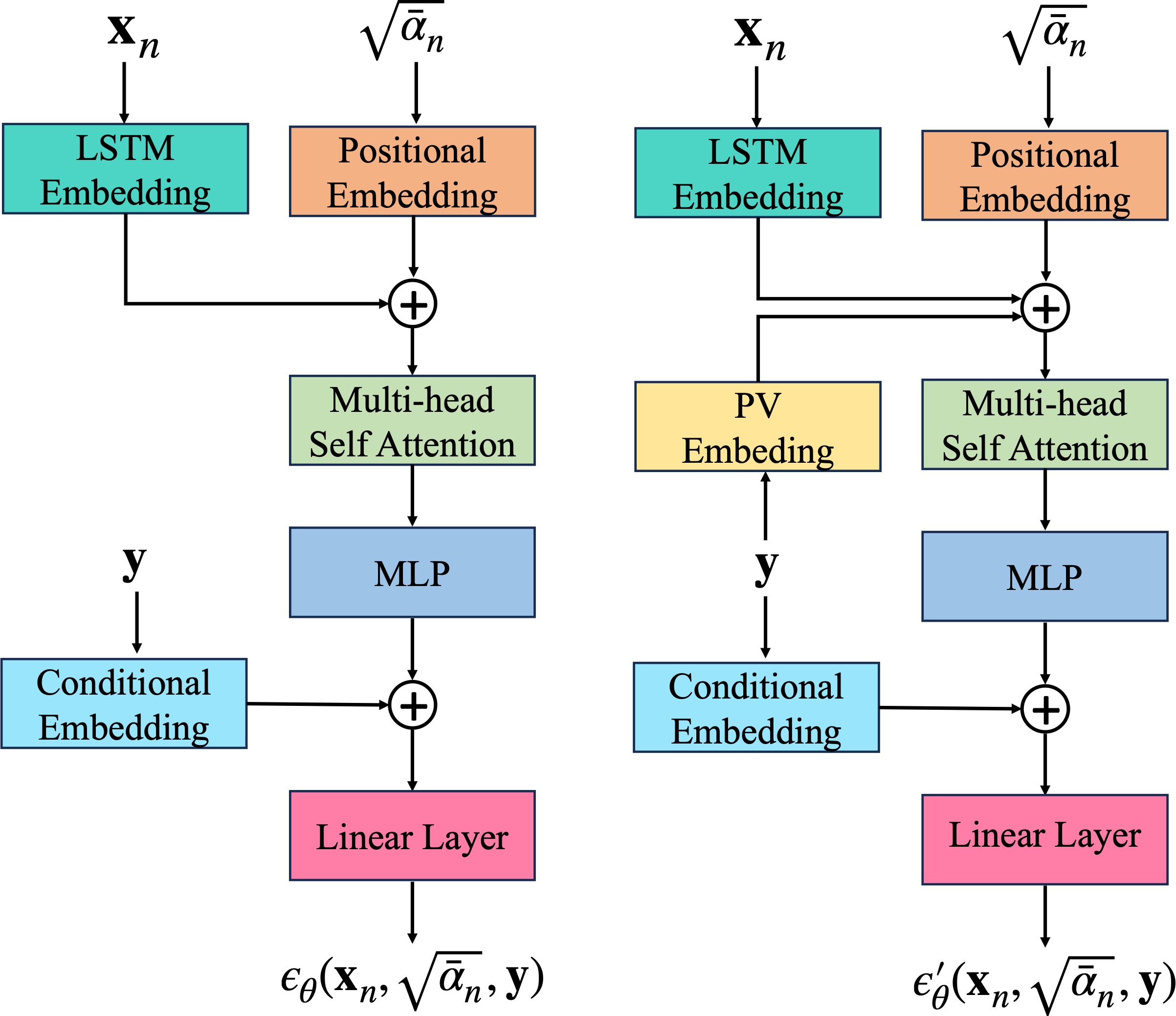}
    \caption{The overall architectures of the baseline denoising networks (left) and physics-informed denoising networks (right).}
    \label{fig:struc}
\end{figure}

The overall architectures of both the baseline and the physics-informed denoising networks are illustrated in Fig. \ref{fig:struc}. The architecture of both the baseline and the physics-informed denoising networks, as outlined, integrates advanced neural network components to effectively handle the complex temporal correlations in the net load time series data. The proposed PDM's network design encapsulates a blend of LSTM networks, multi-head self-attention mechanisms, and a solar PV embedding module to enhance the model's accuracy and efficiency. The structure and operational mechanism of these components are presented below.

\textbf{LSTM Embedding}: Central to capturing temporal correlations in net load profiles, the LSTM network offers a robust framework for modeling time-dependent data. By utilizing a 1-layer LSTM network to process the Gaussian noise input $\mathbf{x}_n \in \mathbb{R}^T$, the model generates latent states that encapsulate temporal patterns and dependencies. The dimension of the LSTM's hidden layer, denoted by $H$, is a crucial hyperparameter, establishing the foundation for the level of temporal complexity attainable by the network.

\textbf{Positional Embedding}: Regulating the diffusion/denoising process's noise level, $\sqrt{\bar{\alpha}_n}$, the Positional Embedding module employs a Transformer-style sinusoidal embedding function. This approach effectively encodes the noise level within the model, ensuring that the temporal dynamics influenced by the diffusion process are accurately represented. The output of this module is adjusted to match the LSTM's hidden layer dimension $H$.

\textbf{Multi-head Self Attention}: The incorporation of a multi-head self-attention module, functioning as a Transformer encoder, introduces a sophisticated mechanism for analyzing and integrating information across different segments of the input data. Maintaining both input and output dimensions at $H$, this module enhances the model's ability to discern complex interdependencies within the net-load data.

\textbf{MLP}: Following the multi-head self-attention module, a Multi-Layer Perceptron (MLP) module with Leaky ReLU activation function in introduced. This module retains the dimensionality at $H$, which helps maintain the processed data's depth and complexity.

\textbf{Conditional Embedding}: This module is another MLP, which encodes the conditional information with dimension $C$ and produces an output with dimension $H$ that aligns with the output of the primary MLP module.

\textbf{Linear Layer}: Concluding the network architecture, a Linear Layer module performs a crucial transformation, which changes the data dimension from $H$ to the final output dimension $T$. This step is pivotal in aligning the processed data with the original signal space, ensuring that the output closely matches the targeted net-load profile.


\begin{figure}[ht]
    \centering
    \includegraphics[width = 0.45\textwidth]{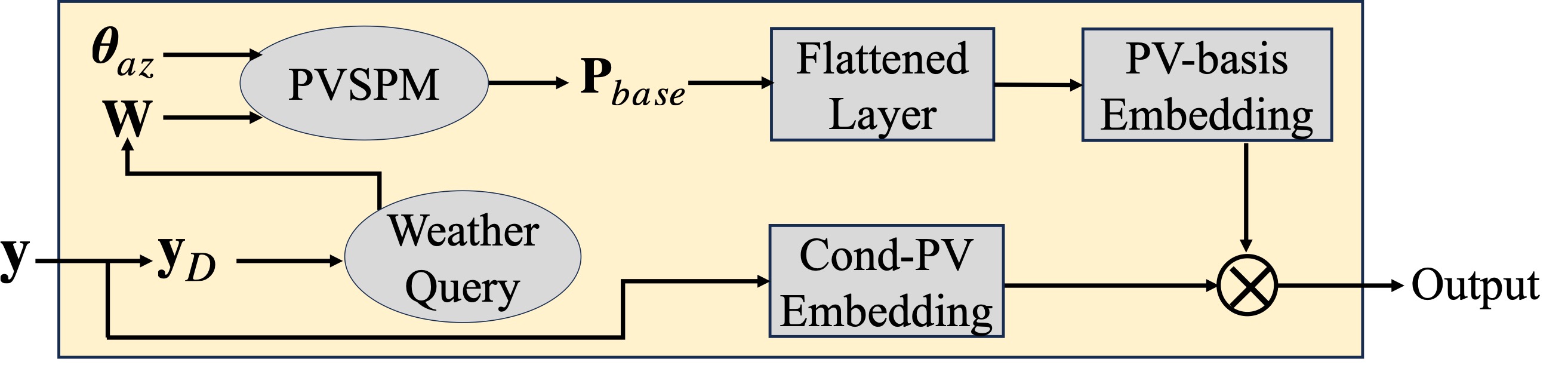}
    \caption{PV embedding architecture. The input of this module is $\mathbf{y}$. $\boldsymbol{\theta}_{az}$ is the base azimuth angle. PV-basis Embedding and Cond-PV Embedding submodules are parameterized by neural networks.}
    \label{fig:PV_emb}
\end{figure}

The details of the PV embedding module are shown in Fig. \ref{fig:PV_emb}. This module contains 5 submodules. The operational flow and functionality of each submodule are outlined below:

\textbf{Weather Query}: This initial step involves taking specific date information, denoted as $\mathbf{y}_D$, and querying to retrieve corresponding weather data. The outcome is a comprehensive set of weather variables for the given date, symbolized as $\mathbf{W}$, which includes temperature, wind speed, DNI, DHI and GHI.

\textbf{PVSPM}: Leveraging the weather data acquired from the previous step, along with representative azimuth angles $\theta_{az}$, this submodule employs the PVSPM to calculate the solar PV generation basis profiles. The output, $\mathbf{P}_{basis}$, is a matrix in $\mathbb{R}^{L \times T}$, where $L$ represents the number of representative azimuth angles, and $T$ signifies the time dimension.

\textbf{Flattened Layer}: Upon receiving $\mathbf{P}_{basis}$, this layer acts to reshape the matrix into a flattened vector. The transformation process effectively converts the $L \times T$ matrix into a one-dimensional vector, facilitating its subsequent processing. This step is crucial for aligning the data structure with the needs of subsequent neural network modules.

\textbf{PV-basis Embedding}: Following the flattening process, this submodule employs a MLP architecture with three layers. It utilizes the Tanh function as its activation mechanism, processing the flattened vector to generate a feature representation that captures the intricacies of solar generation potential under varying weather conditions.

\textbf{Cond-PV Embedding}: Parallel to the PV-basis Embedding, this submodule is another MLP, albeit with two layers. It also adopts the Tanh function for activation. The primary role of this submodule is to further process and refine the information, preparing it for final integration into the diffusion model.

The culmination of these steps is the element-wise multiplication ($\otimes$) of the outputs from the PV-basis Embedding and Cond-PV Embedding submodules. This operation fuses the learned representations of solar generation and other conditional information, resulting in a highly nuanced feature vector ready to be utilized within the diffusion model's framework. This integrated approach allows the model to leverage detailed solar PV generation information, thereby enhancing its predictive accuracy and relevance for applications involving physical signals.
Finally, the input size and output size for each module are summarized in Table \ref{tab: size}.

\begin{table}[ht]
\caption{The Input and Output Size of Each Module}
\centering
\begin{tabular}{ccc}
\hline
\hline
Module & Input size & Output size \\
\hline
LSTM Embedding & $(B, T)$ & $(B, H)$ \\
Positional Embedding & $(B, 1)$ & $(B, H)$ \\
Multi-head Self Attention & $(B, H)$ & $(B, H)$ \\
MLP & $(B, H)$ & $(B, H)$ \\
Conditional Embedding & $(B, C)$ & $(B, H)$ \\
PV Embedding & $(B, C)$ & $(B, H)$ \\

PVSPM & $(B, 5, T)$ & $(B, L, T)$\\
Flattened layer & $(B, L, T)$ & $(B, L \times T)$ \\
Cond-PV Embedding & $(B, C)$ & $(B, H)$ \\
PV-basis Embedding & $(B, L \times T)$ & $(B, H)$ \\
Linear Layer & $(B, H)$ & $(B, T)$ \\
\hline
\end{tabular}
\label{tab: size}
    
\end{table}

\section{Neumerical Studies}
\label{sec: nenumerical_studies}
\subsection{Dataset and Preprocessing}

\textbf{Net load data}. The net load data for residential customers in Austin, Texas, collected by Pecan Street Inc. are used in our experiments \cite{PecanStreet2020}. This dataset includes the daily energy consumption, solar PV production, and net-load time series of individual customers. Specifically, the net load data of 25 residential households from January 1, 2018, to December 30, 2018 are provided. Measurements were recorded at 15-minute intervals, achieving a 99\% completeness rate across all intervals for the 25 homes. The missing values are imputed using the average net load of the adjacent hours. Given the 15-minute sampling rate, the dimensionality of the daily net load profile, $T$, equals $96$. Furthermore, the net load profiles for each customer are Min-max normalized to fall within the range from $-1$ to $1$.

\textbf{Conditional information}. The conditional information $\mathbf{y}$ contains user ID, solar PV system information, and variables associated with the specific dates. For user identification, We employ one-hot encoding for the 25 users, denoted as $\mathbf{y}_{ID} \in \mathbb{R}^{25}$. The dataset provides detailed solar PV system information, including the total and orientation-specific (west, south, east) capacities of solar PV systems installed at each household, represented as $\mathbf{y}_{PV} \in \mathbb{R}^4$. To account for the variations in electricity usage behavior across time, we represent specific dates using the following variables: one-hot encoding for month (12 dimensions), day of the week (7 dimensions), and day of the month (31 dimensions), resulting in $\mathbf{y}_{date} \in \mathbb{R}^{50}$. Consequently, the total dimension of the conditional information, $\mathbf{y} = [\mathbf{y}_{ID}, \mathbf{y}_{PV}, \mathbf{y}_{date}]^T$ is $79$.

\textbf{Weather data}. Hourly weather variables such as DHI, GHI, DNI, temperature and wind speed, are collected from the National Solar Radiation Database \cite{sengupta2018national}. The meteological variables are then converted to 15-minute intervals through linear interpolation to align with the net load data. The approximate longitude and latitude of Austin, Texas $(30.27^{\circ} \text{N}, -97,74^{\circ} \text{E})$ is used as a common proxy location for all customers as their exact locations are not available. The tilt angle $\theta_t$ used to calculate the solar PV generation basis profiles is set as the latitude, i.e., $\theta_t = 30.2672$. The azimuth angle is set from $90^\circ$ to $270^\circ$ with interval of $30^\circ$. Then we can generate AC output for panels facing different directions ranging from east to west.

In the experiments, we randomly select net load profiles for $5$ customers with $365$ days, thus the total number of dataset contains $1825$ trajectories. For each customer, we randomly select $60 \%$ of the dataset for training and the remaining dataset are used for testing.

\subsection{Experimental Setup}

Our benchmark models include WGAN, VAE and NF developed in \cite{dumas2022deep}. The idea behind GANs are adversarial training of two neural networks: the generator and the discriminator. The state-of-the-art Wasserstein GAN with gradient penalty (WGAN-GP) limits the gradient norm of the discriminator’s output with respect to its input to enforce the 1-Lipschitz conditions \cite{gulrajani2017improved}, which can reduce mode collapse and improve the stability of training. VAE contains an encoder and a decoder network, which are jointly trained to maximize a lower bound on the likelihood. NF learns a sequence of transformations, a ﬂow, from a density known analytically, e.g., a normal distribution to a complex target distribution \cite{dumas2022deep}.

Our study also uses a baseline diffusion model for benchmarking purposes. To ensure a fair evaluation, both the baseline diffusion model and the physics-informed model were configured with identical hyperparameters. To balance sample generation quality and speed, we design the noise schedule, $\beta_t$, as a linear function for $t = 1, \cdots, 500$. To accelerate training, we employ a learning rate scheduler that applies a decay factor of 0.9 to the learning rate every 1000 steps. During the diffusion process, the result of each iteration is clipped to be within the range from $-1$ to $1$. Additionally, we utilize an exponential moving average (EMA) for parameter updates in our model. The EMA model is refined at each step by combining $\mu$ times the current EMA model with $1 - \mu$ times the newly updated weights from the most recent forward and backward pass. Here $\mu$ is called the ``smoothing factor".

All the hyperparameters of diffusion models are shown in the Table \ref{tab:Setup_NN}.
Training is completed on a Linux machine with Nvidia 2080 Ti GPU. It's worth noting that the training time for diffusion models exceeds 2 hours, and the sampling time for 1 trajectory is about 15 s. Note that we plan to reduce the training and sampling time in the future work. In this paper we mainly focus on improving the quality of the synthetic data.

\begin{table}[htbp]
    \centering
    \caption{Hyperparameters of the diffusion models}
    \begin{tabular}{cc}
    \hline
    \hline
    Hyperparameter & Value \\
    \hline
    Hidden dimension ($H$) & 1000 \\
    Learning rate & $5 \times 16^{-4}$  \\
     Epoch & $80000$ \\
     EMA decay rate & $0.9$ \\
     Batch size & $5000$ \\
     $\beta_1$ & $10^{-6}$ \\
     $\beta_N$ &  $2 \times 10 ^{-2}$ \\
     Diffusion steps ($N$) & $500$ \\
     Number of heads in attention layer & $4$ \\
     smooting factor $\mu$ & $0.9$ \\
     
     \hline
    \end{tabular}
    \label{tab:Setup_NN}
\end{table}


\subsection{Result and Analysis}


For evaluating synthetic images, the key metrics are the Fr\'echet Inception Distance (FID) and Inception Score (IS). These metrics have been widely applied to assess the quality and diversity of the images produced by generative models, providing insights into how well a model can generate new, realistic images that resemble a given dataset. However, the evaluation metrics developed for synthetic image can not be directly applied for net-load profiles creation. This is because the number samples for net-load profiles is much smaller than that of the image datasets. Thus, we introduce an evaluation system with a set of complementary metrics to thoroughly assess the quality and diversity of synthetic net load data. Our evaluation framework is structured into four parts:

\textbf{Synthetic Conditional Net Load Data}. This part of the evaluation compares synthetic net load profiles generated under given conditions with actual data. The objective is to evaluate the generative models' ability to capture complex net load patterns accurately.

\textbf{t-SNE Visualizations for Each Customer's Net Load Profile}. Utilizing t-SNE visualizations for each customer, this part of the evaluation aims to uncover the generative models' ability to reproduce the nuanced conditional distributions that are unique characteristic of individual customers. It provides insight into whether the generative models can learn the diverse and specific net-load patterns of different users.

\textbf{Probabilistic Forecasting of Marginal Distribution}. In this part, continuous ranked probability score (CRPS) and quantile score (QS) are employed to assess the accuracy of probabilistic forecasts. CRPS evaluates the models' skill in forecasting marginal distributions for specified times of the day. QS is used to compare the accuracy of forecasts at specific quantiles, highlighting the models' precision in predicting extreme conditions.

\textbf{Other Quantitative Metrics}. In this part of the evaluation, we select 6 quantitative metrics to measure the overall similarity between the distribution of the generated and actual data. 


\subsubsection{Synthetic Conditional Net Load Data}
\begin{figure*}
    \centering
    \includegraphics[width = 0.8 \textwidth]{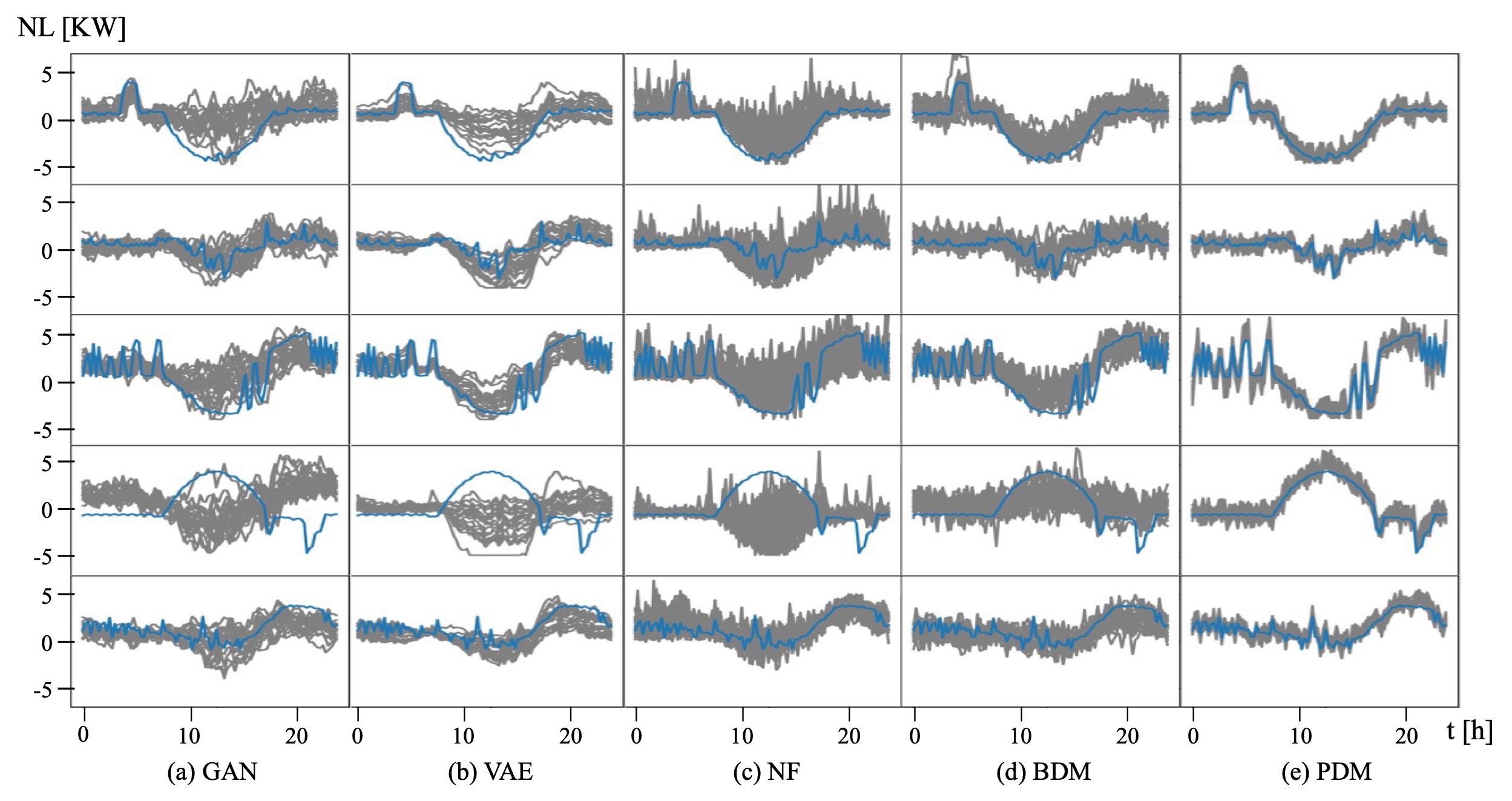}
    \caption{Synthetic net load profiles given 5 sets of conditional information. Each row corresponds to a unique condition, while different columns refer to different generative models. We select conditional information with $5$ different net load patterns to showcase the capability of the generative models. The blue curve represent the actual net load profile, while the gray curves are the 20 synthetic net load profiles.}
    \label{fig:1day}
\end{figure*}

The net load profiles' complex patterns shift significantly based on the conditional information, which includes user ID, solar PV system information, and date. Not all generative models can capture the highly nonlinear relationships between the conditional information and the net load time series. To evaluate the generative models' response to different conditions, we first select 5 typical conditions from the test data set, which are associated with 5 different net load patterns. Next, under each condition, we generate $20$ samples for all of the baseline and proposed generative models. The synthetic net-load data are illustrated in Fig. \ref{fig:1day}. For every condition, the PDM yields the best performance. In contrast, the GAN, VAE, NF model were unable to accurately replicate diverse net load patterns. While the BDM and the proposed PDM showed great ability to emulate each pattern, the BDM exhibited higher approximation error and variance compared to the proposed PDM.

The approximation error of a sample created by generative models can be measured by mean squared error (MSE). The MSE of the synthetic net-load data for a given condition $\mathbf{y}$ can be calculated as:
\begin{equation}
    MSE(t \vert \mathbf{y}) =  \sum_{i=1}^{M} (\hat{x}_{y, t}^{i} - x_{\mathbf{y}, t}^{i} ) ^2,
\end{equation}
where $M$ represents the number of samples. $\hat{x}_{\mathbf{y}, t}^{i}$ denotes the predicted value of the generative model and $x_{\mathbf{y}, t}^{i}$ indicates the actual value at time $t$ for the $i$-th sample under condition $\mathbf{y}$.

\begin{figure}
    \centering
    \includegraphics[width = 0.5 \textwidth]{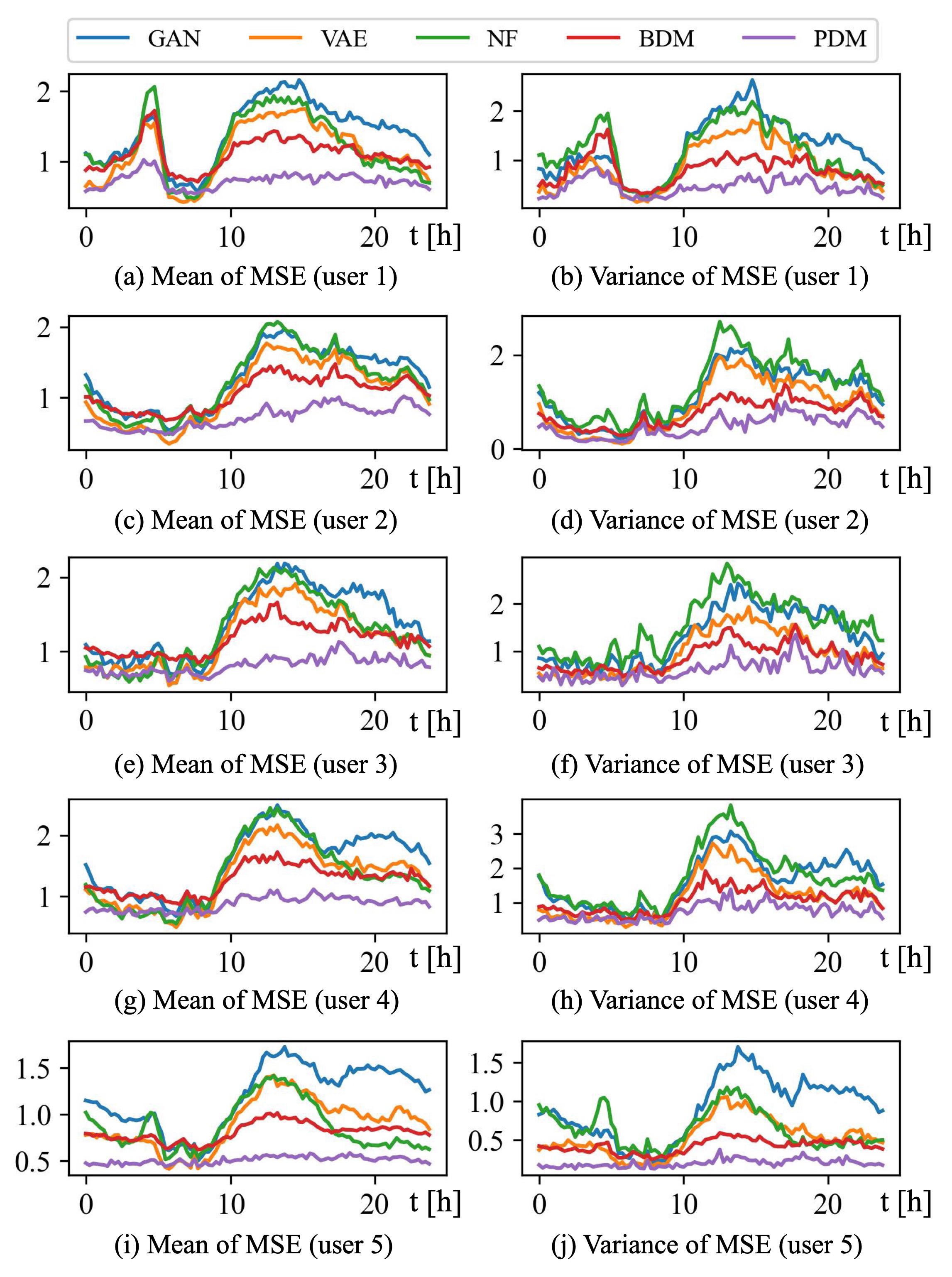}
    \caption{Mean and variance of the MSEs for each customer's net load profiles.}
    \label{fig:mv}
\end{figure}

For each customer, we calculate the MSEs between the predicted and actual net load profiles over time across all 5 conditions. Subsequently, we determine the mean and variance of their respective MSEs. The results are presented in Fig. \ref{fig:mv}. The figure reveals that for each customer, the proposed PDM achieves the lowest mean and variance for nearly all time points, while the BDM achieves the second best result. The results show that the proposed PDM is much more capable than state-of-the-art generative models in learning the conditional distribution of the net load data, particularly in scenarios that do not occur frequently.

\subsubsection{t-SNE Visualizations for Each Customer's Net Load Profile}
\begin{figure}
    \centering
    \includegraphics[width = 0.45 \textwidth]{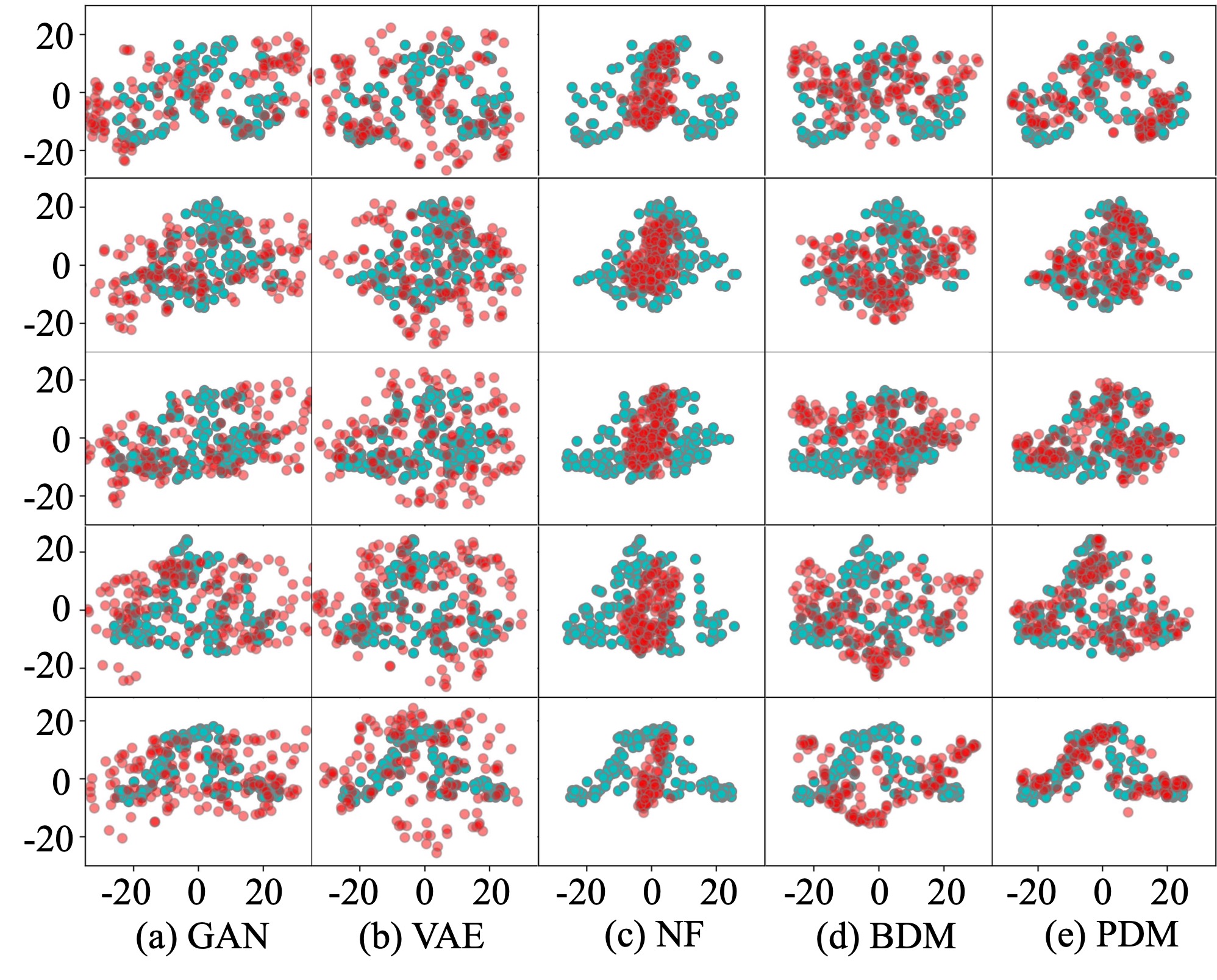}
    \caption{t-SNE visualizations. Each dot represents a daily net load profile. Green and red represent real and synthetic data respectively, where a greater overlap of green and red dots shows a higher distributional-similarity between the generated and the actual data.}
    \label{fig:tnse}
\end{figure}

We employ t-SNE to project the high-dimensional net load data to lower dimensional space to explore the ability of generative models to capture the nuanced conditional distributions of individual customers' net load profiles, which are influenced by their unique electricity consumption habits and solar PV system configurations. The t-SNE visualizations shown in Fig. \ref{fig:tnse} illustrate the degree of agreement between the distributions of the generated synthetic data and the actual net load data for specific customers. 

As shown in the figure, while GAN and VAE models sometimes create data points that look like the real data in the low dimensional space, their performance is marred by inconsistencies. Although capable of approximating the overarching structure of the data, these two models falter in replicating intricate patterns. In contrast, the NF model commits fewer errors in generating data closely aligned with actual observations. Nevertheless, they fall short in representing the entire spectrum of patterns in the dataset, indicating a somewhat limited representation power. Both the BDM and the PDM demonstrate a balanced ability to learn data distributions, effectively avoiding the pitfalls of being either overly conservative or excessively aggressive in data generation. Notably, the proposed PDM stands out for its exceptional ability to accurately model the conditional distributions, marking it the superior choice among all evaluated models.

\subsubsection{Probabilistic Forecasting of Marginal Distribution}

\begin{figure}
    \centering
    \includegraphics[width = 0.45 \textwidth]{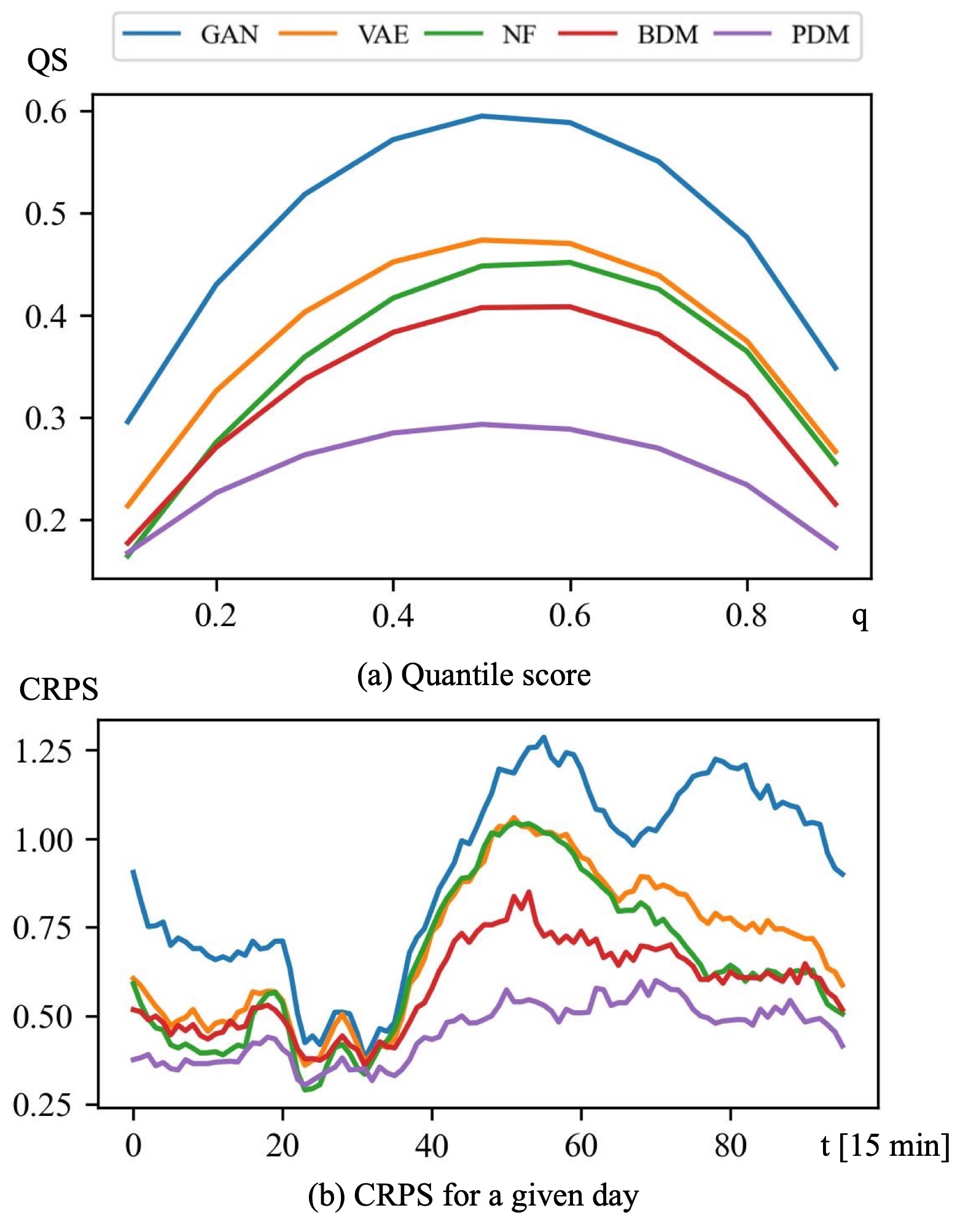}
    \caption{QS and CRPS per marginal. Quantile score (a): the lower and the more symmetrical, the better. CRPS of time slots (b): the lower, the better.}
    \label{fig:PS_CRPS}
\end{figure}

The quantile score (QS) and continuous ranked probability score (CRPS) are used as evaluation metrics for the performance of probabilistic forecasting of marginal distribution \cite{dumas2022deep}. We calculate the QS for different level of quantile forecasts. The quantile score is plotted against different quantiles ($q = [0.1, 0.2, ..., 0.9]$) in Fig. \ref{fig:PS_CRPS}(a). A lower score indicates that the predicted quantiles are closer to the true quantiles of the actual data. The proposed PDM achieves the lowest quantile scores across all quantiles and it is shown to excel at predicting true quantiles of the net load profile distribution. 

CRPS takes into account both the accuracy and the sharpness of the predictions. This means that it not only measures how close the predictions are to the actual values but also how confident the predictions are (i.e., a narrower predicted distribution is better if it is accurate). The CRPS for all time slots of a given day are shown in Fig. \ref{fig:PS_CRPS}(b). The PDM achieves the lowest level for nearly all time slots. The result demonstrates the superior performance of PDM in probabilistic forecasting.

\subsubsection{Quantitative Metrics Evaluation}

Multiple quantitative metrics are used to evaluate the quality of synthetic net load data, including: MAE, root mean squared error (RMSE), averaged quantile score ($\overline{\text{QS}}$), averaged continuous ranked probability score ($\overline{\text{CRPS}}$), energy score (ES) and variogram score (VS) \cite{dumas2022deep}. The MAE and RMSE metrics are pivotal in evaluating the accuracy of generated results under specific conditions. The averaged QS enables detailed assessment of forecast quality at specific probability levels, such as over-forecasting or under-forecasting, particularly with regard to the tails of the predictive distribution \cite{lauret2019verification}. The ES and VS are used as evaluation metrics for multivariate data generation \cite{scheuerer2015variogram}. In this paper, we consider the trajectory of net load in a day as a vector of multivariate variables. For the calculation of VS, we use equal weights across all hours of the day and setting $\gamma$ at 0.5 \cite{dumas2022deep}.

The results for the 6 quantitative metrics are presented in Table \ref{tab:eva_metric}. These metrics indicate that BDM significantly outperforms GAN, VAE, and NF across all six metrics. Moreover, PDM surpasses BDM in every metric, showcasing its superior performance. The improvement of PDM for all the metrics exceeds $20 \%$. The numerical study results suggest that the proposed PDM excels in modeling the complex patterns and unique distribution of net load profiles. Its effectiveness can be attributed to the incorporation of domain knowledge through a physics-informed model. The BDM although not as powerful as the PDM, also demonstrates superior performance over state-of-the-art generative models, which implies that the diffusion model have an advantage over other generative models in the task of net-load profile generation.

\begin{table}[ht]
\centering
\caption{Quantitative Evaluation Metrics}
\begin{tabular}{ccccccc}
\hline
\hline
\renewcommand{\arraystretch}{2}
Model & MAE & RMSE & \rule{0pt}{2.5ex} $\overline{\text{QS}}$ & $\overline{\text{CRPS}}$ & ES & VS \\

\hline
WGAN & 1.38 & 3.35 & 0.49 & 0.90 & 11.07 & 1504.04 \\
VAE & 1.13 & 2.35 & 0.38 & 0.70 & 8.89 & 1280.41 \\
NF & 1.20 & 2.94 & 0.35 & 0.64 & 8.69 & 1303.79 \\
BDM & 1.08 & 2.01 & 0.32 & 0.58 & 7.22 & 1078.53 \\
PDM & \textbf{0.73} & \textbf{1.05} & \textbf{0.24} & \textbf{0.45} & \textbf{5.72} & \textbf{803.24} \\

\hline
\end{tabular}
\label{tab:eva_metric}
\end{table}

\section{Conclusion}
\label{sec:conclusion}
In this paper, we propose a novel physics-informed diffusion model for generating synthetic net load data by embedding solar PV system performance model into the baseline diffusion model. To mitigate the convergence problem in the model training process, we propose to jointly learn the parameters of the diffusion model for load and the physics-based model for solar PV system. A unique denoising neural network with conditioning is designed to generate the net load data for different customers and weather conditions. Comprehensive numerical results on Pecan Street dataset reveals that the proposed PDM significantly outperforms other generative models such as GANs, VAEs, NFs and BDM. Specifically, the proposed PDM yields more than 20\% improvement across all evaluation metrics compared to state-of-the-art generative models. Future extensions of this work include scaling the model for larger datasets and considering other types of behind-the-meter resources.

\bibliographystyle{IEEEtran}
\bibliography{IEEEabrv, refs}
\end{document}